# A hybrid Bayesian network for medical device risk assessment and management

Joshua L. Hunte[a]*, Martin Neil[a,b] and Norman E. Fenton[a,b]

[a]Risk and Information Management Research Group, School of Electronic Engineering and Computer Science, Queen Mary University of London, London, E1 4NS, UK; [b]Agena Ltd, Cambridge, UK

*Corresponding author address: Joshua Hunte, Risk and Information Management Research Group, School of Electronic Engineering and Computer Science, Queen Mary University of London, London, E1 4NS, UK. email address: j.l.hunte@qmul.ac.uk

**Abstract**

ISO 14971 is the primary standard used for medical device risk management. While it specifies the requirements for medical device risk management, it does not specify a particular method for performing risk management. Hence, medical device manufacturers are free to develop or use any appropriate methods for managing the risk of medical devices. The most commonly used methods, such as Fault Tree Analysis (FTA), are unable to provide a reasonable basis for computing risk estimates when there are limited or no historical data available or where there is second-order uncertainty about the data. In this paper, we present a novel method for medical device risk management using hybrid Bayesian networks (BNs) that resolves the limitations of classical methods such as FTA and incorporates relevant factors affecting the risk of medical devices. The proposed BN method is generic but can be instantiated on a system-by-system basis, and we apply it to a Defibrillator device to demonstrate the process involved for medical device risk management during production and post-production. The example is validated against real-world data.

# 1. Introduction

The medical device industry requires that devices used by patients and healthcare professionals are acceptably safe. There are several standards for medical device safety, such as IEC 60601-1 (IEC, 2015a), but ISO 14971 (ISO, 2019) is the primary standard used by medical device manufacturers. In fact, other standards for medical device safety make normative references to ISO 14971. This standard provides a framework for medical device manufacturers to manage the risks associated with medical devices throughout their life cycle (i.e., initial conception to final decommissioning and disposal). It specifies a set of requirements and expectations for medical device risk management. For instance, ISO 14971 includes requirements for risk analysis (i.e., hazard identification and risk estimation), risk evaluation, risk control and evaluation of overall residual risk. However, ISO 14971 does not specify a particular method or process for medical device risk management. Hence, the methods used for medical device risk management by medical device manufacturers may vary due to the type of medical device and available information (e.g., testing data) and may require validation. In particular, there are several risk analysis methods for medical devices, including the commonly used Fault Tree Analysis (FTA) and Failure Modes and Effects Analysis (FMEA) (ISO, 2020). However, these classical risk analysis methods have limitations such as: limited handling of dependencies among system components; handling uncertainty in testing and failure data; providing reasonable risk estimates with limited or no testing and failure data; and incorporating expert judgement. These limitations are resolved using Bayesian networks (BNs) (Bobbio, Portinale, Minichino, & Ciancamerla, 2001; Fenton & Neil, 2018; Hunte, Neil, & Fenton, 2022a; Rausand & Hoyland, 2003; Weber, Medina-Oliva, Simon, & Iung, 2012).

In this paper, we propose a novel systematic method for medical device risk assessment and management using hybrid Bayesian networks (BNs) that: improves the handling of uncertainty; uses causal knowledge of the risk management process; incorporates relevant factors affecting the safety and risk of medical devices; complements existing medical device risk management tools and methods; uses quantitative data and expert judgement. Bayesian networks (BNs) are suitable for medical device risk management due to their ability to handle uncertainty and produce results using objective and subjective evidence (Fenton & Neil, 2018; Pearl & Mackenzie, 2018). Also, they are used for risk assessment in several domains, including systems reliability, health, railway, finance and consumer product safety (Berchialla, Scarinzi, Snidero, & Gregori, 2010; Hunte et al., 2022a; Leu & Chang, 2013; Li, Liu, Li, & Liu, 2019; Marsh & Bearfield, 2004; Neil, Marquez, & Fenton, 2008; Weber et al., 2012). The proposed generic BN for medical device risk management extends the previous work on using BNs for consumer product safety and risk assessment (Hunte et al., 2022a) and provides a robust systematic method for medical device manufacturers to meet the requirements of ISO 14971 (see Figure 1).

Please note that the main standard referred to throughout this paper is ISO 14971 (ISO, 2019) and its accompanying guidelines for application ISO/TR 24971 (ISO, 2020). Unless other citations are provided, all definitions in this paper refer to this standard and its guidelines.

The rest of this paper is organised as follows: Section 2 provides an overview of medical device risk management and risk analysis methods. Section 3 discusses the limitations of risk analysis methods. Section 4 describes our method, including BN model development. Section 5 presents the results of risk management scenarios. The results are discussed in Section 6, and the limitations of the study in Section 7. Finally, conclusions and recommendations for further work are presented in Section 8.

## 2. Background

### 2.1 Medical device

A *medical device* is "any instrument, apparatus, implement, machine, appliance, implant, reagent for in vitro use, software, material or other similar or related article, intended by the manufacturer to be used, alone or in combination for a medical purpose" (ISO, 2019). There are two main types of medical devices based on use: *single-use* and *multiple-use.* A *single-use medical device* is a medical device "intended to be used on an individual patient during a single procedure and then discarded" (MHRA, 2021). A *multiple-use (reusable) medical device* is a medical device "that health care providers can reprocess and reuse on multiple patients" (FDA, 2018b). Though single-use and multiple-use medical devices may contain software, the software can be considered a medical device on its own (Software as a medical device). *Software as a medical device* (SaMD) is "software intended to be used for one or more medical purposes that perform these purposes without being part of a hardware" (IMDRF, 2013).

### 2.2 Medical device risk management

*Medical device risk management* is the "systematic application of management policies, procedures and practices to the tasks of analysing, evaluating, controlling and monitoring risk" of medical devices (ISO, 2019). The international standard for medical device risk management ISO 14971 requires that manufacturers have a documented process for managing the risks associated with medical devices. It specifies a set of requirements and expectations for the documented risk management process (see Figure 1) applicable to the complete life cycle of the medical device (i.e., initial conception to decommissioning and disposal). However, ISO 14971 does not specify a particular process or method for performing risk management. Therefore, medical device manufacturers are free to develop or use suitable risk management methods and processes, e.g., the BXM method (Elahi, 2022), to satisfy the requirements of ISO 14971. Figure 1 shows ISO 14971 risk management process and requirements.

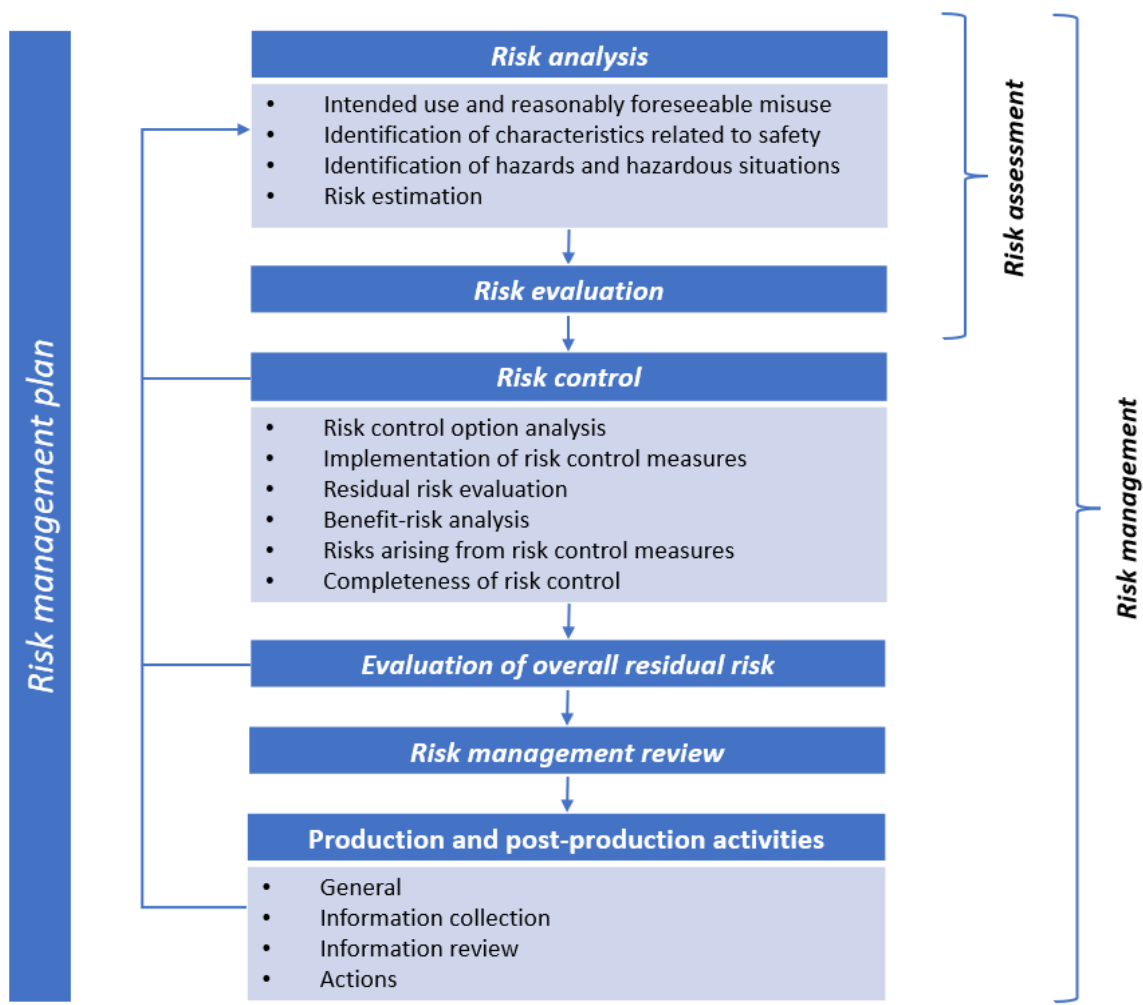

Figure 1. ISO 14971 risk management process

A key component of the ISO 14971 risk management process is risk assessment, which consists of two activities: risk analysis and risk evaluation. During risk analysis, the hazards associated with medical devices are identified, and the risk for each hazard is estimated. A *hazard* is a "potential source of harm". Techniques such as Preliminary Hazard Analysis (PHA) are used to identify hazards associated with medical devices. The schematic shown in Figure 2 provides an overview of ISO 14971 risk estimation. *Risk* is "the combination of the probability of occurrence of harm $P$ and the severity of the harm $S$" i.e., $Risk = P \times S$. The probability of occurrence of harm $P$ is the product of the probability of the hazardous situation occurring $P_1$ and the probability of the hazardous situation causing harm $P_2$, i.e., $P = P_1 \times P_2$. A *hazardous situation* is a "circumstance in which people, property, or the environment are exposed to one or more hazards", such as normal device use. Techniques used to estimate risk include risk matrices and FTA.

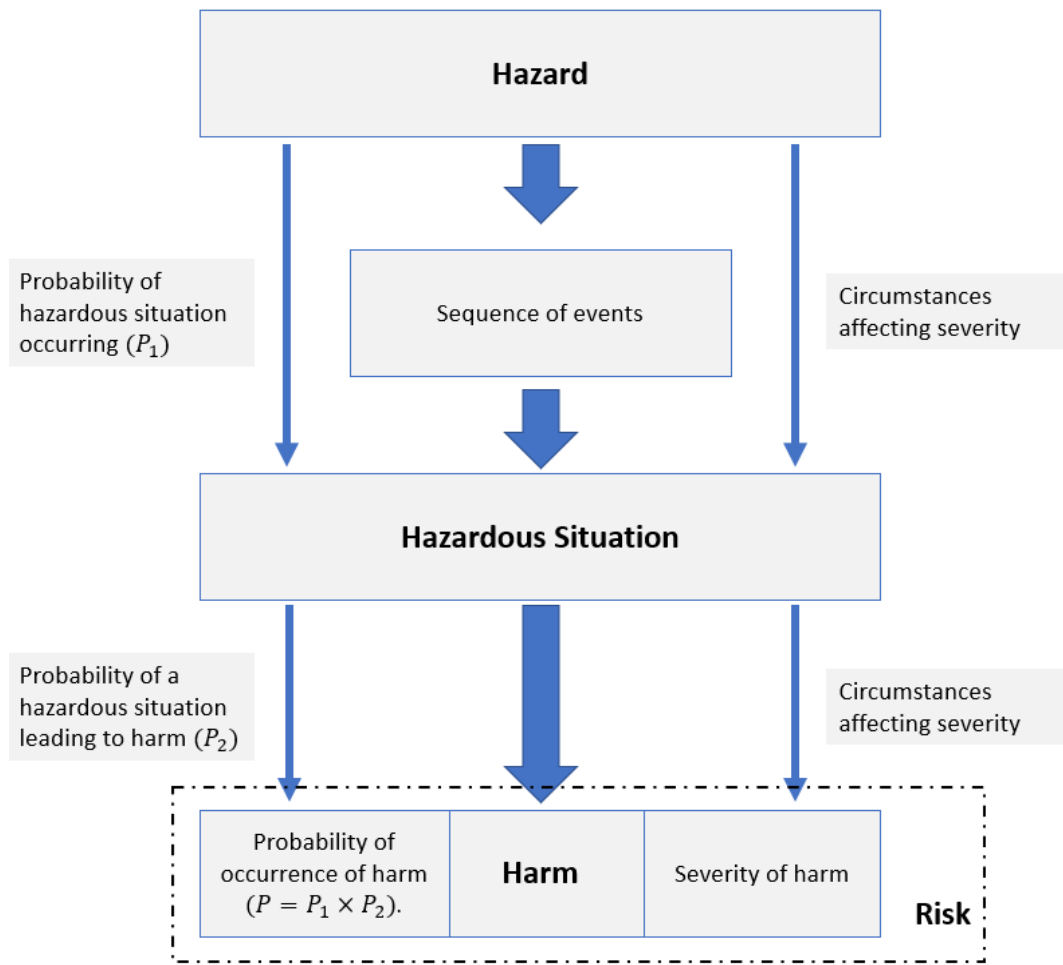

Figure 2. ISO 14971 Risk Estimation.

The estimated risk for each hazard or hazardous situation is then evaluated (i.e., risk evaluation) to determine risk acceptability using the defined criteria for risk acceptability. In situations where the estimated risk is judged not acceptable, additional risk control measures are implemented to reduce the risk to an acceptable level. In situations where risk reduction is not practicable, a benefit-risk analysis is done. However, in situations where the estimated risk is judged acceptable, additional risk controls are not required and the estimated risk is viewed as the *residual risk* (i.e., risk remaining after risk control measures are applied).

## 2.3 Risk Analysis Methods

There are several methods that can be used for risk analysis in the medical device industry (Elahi, 2022; ISO, 2020; Ramakrishna, Tian, Wang, Liao, & Teo, 2015). These methods are complementary and can be used as required to perform a comprehensive risk analysis. In this paper, we review the following commonly used risk analysis methods discussed in ISO/TR 24971: Preliminary Hazard Analysis (PHA), Fault Tree Analysis (FTA) and Failure Mode and Effects Analysis (FMEA).

### 2.3.1 Preliminary Hazard Analysis

A Preliminary Hazard Analysis (PHA) is a technique commonly used during the early stages of the development process when there is little information about the design or operation of a medical device. It is used to identify hazards and hazardous situations that can result in harm for a medical device. The PHA method includes the following steps:

1. Identify and define the scope and objectives of the analysis.
2. Identify and collate relevant information about the medical device, e.g., system requirements and its potential hazards.
3. Identify hazards associated with the medical device.
4. Define suitable hazardous situations and potential harms for each hazard.
5. Determine the probability of occurrence of harm $P$ (see Figure 2).
6. Determine the severity of the harm $S$.
7. Compute the risk of the medical device (see Figure 2).
8. Identify potential risk control measures and compute the residual risk of the medical device given risk control measures.

The results of a PHA can inform future product development activities, risk control measures and other risk analysis techniques. A number of studies have used PHA to identify potential hazards and hazardous situations for medical devices (Elahi, 2022; Masci, Zhang, Jones, Thimbleby, & Curzon, 2014; Torrez, 2018a; Y. Zhang, Jones, & Jetley, 2010). For instance, Zhang et al. (2010) used a PHA to identify hazards and hazardous situations for an insulin infusion pump. Masci et al. (2014) used a PHA to identify hazards for the number entry part of an infusion pump interface. Torrez (2018) used a PHA in combination with an FTA and FMEA to identify hazards and risks that can lead to the failure of an Integra UltraVS Neonate Valve.

## 2.3.2 Fault Tree Analysis

A Fault Tree Analysis (FTA) is a deductive, top-down risk analysis method that starts from an assumed undesired event, e.g., a hazardous situation, followed by the identification of all its causes and contributing factors. The assumed undesired event is called the *top event*, its immediate causes are called *intermediate events,* and its root causes are called *basic events*. An FTA can be used to perform quantitative and qualitative analyses at different stages of the product's life cycle. Quantitative analysis is done by assigning probabilities to the basic events to compute the probability of the top event using probability rules. The results of the FTA are presented as a fault tree (FT). A *fault tree* is a graphical model that describes the logical relationship between top events, intermediate events and basic events using logic gates. It provides useful insights, such as the set of basic events (called a cut set) that can cause the top event to occur.

The results of an FTA can inform and complement other risk analysis methods such as FMEA. A number of studies have used FTA for medical device risk analysis (Hyman, 2002; Hyman & Johnson, 2008; Rice, 2007). For instance, Hyman (2002) used it to analyse use errors for medical devices. Hyman and Johnson (2008) used it to analyse patient harm associated with clinical alarm failures. Rice (2007) recommended FTA for medical device risk-based evaluation and maintenance.

## 2.3.3 Failure Mode and Effects Analysis

A Failure Mode and Effects Analysis (FMEA) is an inductive, bottom-up risk analysis method used to identify the failure modes of a system and the causes and effects of those failures on the system. A *failure* is the inability of the system to operate as intended. Given a medical device, an FMEA will identify the failure modes, such as hardware failures, that can cause hazards or hazardous situations to occur. It will also analyse the effect of failure modes on the device, its users, and the environment to determine their severity.

The FMEA method includes the following steps:

1. Identify and define the scope and objectives of the analysis, including the failure mode to be investigated and acceptability criteria.
2. Collate and organise relevant resources for the analysis, such as the risk team and the documents, e.g., system requirements.
3. Perform the analysis:

a. Identify the cause and effect of the failure mode.
b. Determine and assign ratings for the Severity $S$ of the effect of the failure (using a 5-point ranking scale ranging from *Negligible* to *Fatal*), the probability of Occurrence $O$ of the failure (using a 5-point ranking scale ranging from *Improbable* to *Frequent*), and the Detection $D$ of the causes of the failure (using a 5-point ranking scale ranging from *Almost certain* to *Undetectable*).
c. Compute the Risk Priority Number (RPN) and Risk. The RPN combines the Severity, Occurrence and Detectability to determine the criticality of the failure mode, i.e., $RPN = S \times O \times D$. The failure mode with the largest RPN is the most critical for the device. The risk of the failure mode is computed using a risk matrix that combines the Severity and Occurrence ratings.
d. Evaluate the acceptability of the RPN and risk given the defined acceptability criteria.
4. Determine appropriate risk control measures given the criticality of the failure mode.

The results of the FMEA can inform future product development activities, risk control measures and other risk analysis techniques. A number of studies have used FMEA for medical device risk management (Clemente et al., 2019; Elahi, 2022; Onofrio, Piccagli, & Segato, 2015; Ramakrishna et al., 2015). However, note that FMEA can also be used to identify failures in the manufacturing process (Process FMEA) and failures associated with device use or misuse (Use FMEA).

## 3. Limitations of Risk Analysis Methods

The above methods have the following limitations when used for medical device risk assessment and management (Bobbio et al., 2001; Eagles & Wu, 2014; Fenton & Neil, 2018; Kabir, 2017; Khakzad, Khan, & Amyotte, 2011; Ruijters & Stoelinga, 2015; Spreafico, Russo, & Rizzi, 2017):

1. *Limited approach to handling uncertainty:* When quantitative data are available for analysis, classical methods use point values instead of distributions to define the probability of events, e.g., failures or hazards. For this reason, they are unable to fully handle uncertainty in the assigned probabilities values (i.e., second-order uncertainty).

2. *Failure to incorporate the causal nature of risk*: Risk is computed as the combination of the probability of occurrence of harm $P$ and the severity of the harm $S$, i.e., $Risk =$ $P \times S$. This method of risk computation does not consider the causal nature of risk, i.e., $P$ and $S$ are dependent on probabilistic information about trigger events (i.e., initiating events that cause the risk event), control and mitigating events (i.e., events that can stop the occurrence of the risk event or mitigate the consequence of the risk event) (Fenton & Neil, 2018). Since the assumptions that the risk is conditioned on are not explicit, this makes the notion of both $P$ and $S$ ill-defined and overly subjective. Furthermore, the 'risk register' type approaches used to compute risk can result in manufacturers who consider all known risks being penalised as compared to those who do not consider all known risks since the former will have higher probabilities of risk.

3. *Cannot compute risk for novel products with limited or no historical data*: Since risk is computed as $P \times S$, classical methods will not be able to provide reasonable risk estimates for novel products or products with limited testing data since the probability of occurrence of harm $P$ may be unknown.

4. *Limited approach to handling multi-state variables:* In situations where a variable or event has more than two states (multi-state), classical methods such as FTA are not suitable for performing analysis since they can only support binary variables (i.e., variables with only two states).

5. *Handling sequence-dependent variables:* In situations where component failures and hazards are causally dependent, classical methods are not suitable since they cannot handle sequence-dependent failures or hazards.

6. *Limited approach to updating prior data/ results given new data*: Revising the results of an analysis given new data, e.g., probability of a basic event occurring, entails repeating the analysis with the new data, which is not practical since it is usually time-consuming and expensive.

7. *Limited approach to combining subjective and objective evidence:* Existing methods may find it difficult to combine objective and subjective evidence to estimate risk.

Some extensions to the existing methods, such as Dynamic FT (Fenton & Neil, 2018; Ruijters & Stoelinga, 2015) have addressed some of these limitations. In this paper, we show that hybrid

Bayesian networks (Fenton & Neil, 2018; Koller & Friedman, 2009) address all of the limitations.

## 4. Method

### 4.1 Bayesian Networks

Bayesian networks (BNs) are directed acyclic graphs consisting of qualitative and quantitative components (Fenton & Neil, 2018; Pearl & Mackenzie, 2018; Spohn, 2008). The qualitative component consists of nodes representing random variables (discrete and continuous) and directed arcs representing the causal relationship between the random variables. For instance, in the BN shown in Figure 3, a directed arc from the node 'Wrong Setting Chosen' to the node 'Surface Too Hot' indicates that the node 'Wrong Setting Chosen' causally influences the node 'Surface Too Hot' or the node 'Surface Too Hot' is dependent on the node 'Wrong Setting Chosen'. In this example, the node 'Wrong Setting Chosen' is called the *parent* node, and the node 'Surface Too Hot' is called the *child* node. The quantitative component of the BN consists of node probability tables (NPTs), describing the conditional probability distribution of nodes given their parents. NPTs can be populated using several methods; for instance, the NPTs for discrete variables (nodes) are defined by specifying the probabilities for each node state or using comparative expressions such as IF statements. The NPTs for continuous or numeric variables (nodes) are defined using mathematical functions and statistical distributions (Fenton & Neil, 2018). For instance, as shown in Table 1, the NPT for the node 'Wrong Setting Chosen' is an Exponential distribution. The NPT for a node without parents (called a root node) is its prior probability distribution. A BN consisting of both discrete and continuous variables (see Figure 3) is called a *hybrid* BN (Koller & Friedman, 2009). Hybrid BNs are useful for modelling scenarios incorporating both continuous and discrete variables, e.g., medical device risk assessment and management.

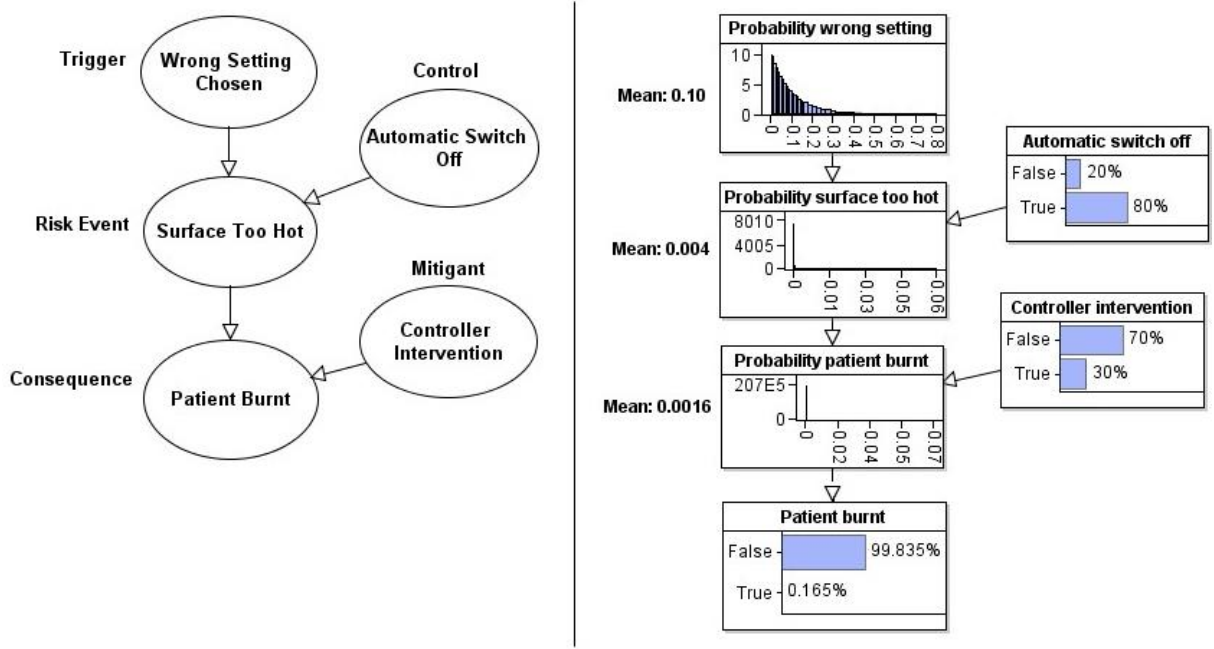

Figure 3. Hot surface risk for a defibrillator BN

Table 1. NPTs for nodes in Hot surface risk for a defibrillator BN

| Node Name | NPT |
| --- | --- |
| **Wrong Setting Chosen** | Exponential (10) |
| **Automatic Switch Off** | False: 0.2, True: 0.8 |
| **Controller Intervention** | False: 0.7, True: 0.3 |
| **Surface Too Hot** | Partitioned expression (False: 0.2 × wrong_setting, True: wrong_setting × 0.001) |
| **Patient Burnt** | Partitioned expression (False: Triangle (0.2 × hot_surface, hot_surface, 0.5 × hot_surface), True: 1.0E-4 × hot_surface) |

Once the structure and the NPTs of a BN are specified, the BN is fully parameterized and can be used for different probabilistic reasonings (or inferences) using Bayes Theorem. Bayes Theorem updates our prior belief of a hypothesis given new evidence. Our prior belief is called the prior probability, and our revised belief is called the posterior probability.

The BN shown in Figure 3 also illustrates the causal context in which risk occurs. In the causal perspective of risk, the risk is characterized by a set of causal interrelated events consisting of the risk event itself, consequence(s), trigger(s), control(s) and mitigating events (Fenton & Neil, 2018). For instance, consider the risk of 'Surface Too Hot' for a defibrillator (see Figure 3), the risk event 'Surface Too Hot' is dependent on the trigger event 'Wrong Setting Chosen' and the control event 'Automatic Switch-Off' stopping the risk event. The consequence event

'Patient burnt' is dependent on the risk event 'Surface Too Hot' and the mitigating event 'Controller Intervention' avoiding the consequence event. In this example, the mean probabilities of the risk and consequence events are 0.004 and 0.0016, respectively, if we assume that the mean probability of the trigger event is 0.1 and the probabilities of the control and mitigating events are 0.80 and 0.30, respectively. In this example, we used discrete nodes to represent control, mitigant and consequence events and continuous nodes to represent trigger, risk and consequence events. We believe that the causal perspective of risk (used to develop the BN for medical device risk assessment and management) supports comprehensive and practical risk analysis since it decomposes a risk problem into a causal chain of events, unlike the classical approach, i.e., $Risk = P \times S$.

There is previous research on using Bayesian networks (BNs) for medical device risk assessment and management. For instance, Haddad et al. (2014) developed a BN to predict fatigue fracture of a cardiac lead. They validated the results of the BN by comparing it to the field performance data for cardiac leads available on the market. Medina et al. (2013) developed a BN to identify the critical factors that affect the decision time for the Food and Drug Administration (FDA) to approve a medical device for market release. Zhang et al. (2016) developed a BN to detect faults associated with medical body sensors network that collects and uses physiological signs for patient health monitoring. Rieger et al. (2011) proposed a Bayesian risk identification model (BRIM) to predict and reduce use error risk during the development of medical devices. Li et al. (2019) used a dynamic BN to assess the risk of device failures and human errors in healthcare. However, the BNs in previous research may not be generalisable (so they cannot be used to assess the risk of different medical devices). In the following section, we present what we believe is the first BN for medical device risk assessment and management that is generalisable, applicable to all stages of the device life cycle, complements classical methods such as FTA and FMEA and supports medical device manufacturers in fulfilling the requirements of ISO 14971.

### 4.2 Methodology for building the hybrid BN

#### 4.2.1 Variables Selection and BN Structure

In this study, a core team of three domain experts identified relevant variables and the initial BN structure using literature and industry experience (Elahi, 2022; IEC, 2015b, 2015a; ISO, 2019, 2020; Ramakrishna et al., 2015; Rausand & Hoyland, 2003). The core components of the BN structure are based on the logical causal nature of how medical devices lead to hazards and then injuries. The BN structure was then presented to two medical device safety consultants for review and feedback.

The structure of the BN was constructed using generic idioms (Neil, Fenton, & Nielsen, 2000) and product safety idioms (Hunte, Neil, & Fenton, 2022b). Idioms are small reusable BN fragments representing generic types of uncertain reasoning (Neil et al., 2000). All model development was done using AgenaRisk Software (Agena Ltd, 2022).

### 4.2.2 Node Probability Tables (NPTs)

The node probability tables (NPTs) for model variables (see Supplemental Table 1) were defined using ranked nodes (Fenton, Neil, & Caballero, 2007), mathematical functions, statistical distributions, and comparative expressions. *Ranked nodes* were used to define discrete variables whose states represent a ranked ordinal scale, e.g., *Benefits of device* node with states: 'low', 'medium', 'high'. A ranked node maps the states of a variable to subintervals of a numerical scale [0,1]. Since ranked nodes use a numerical scale, their NPTs can be defined using statistical distributions. In the BN model, ranked nodes with parents are defined using a TNormal distribution with mean $\mu$ as a weighted function of its parents and variance $\sigma^2$, whereas ranked nodes without parents are defined using a Uniform distribution.

*Mathematical functions* were used to define the NPTs for some continuous (numeric) variables given their parents. For instance, the NPT for the variable *probability of a fatal injury per demand* is the mathematical function: *probability of hazard per demand × probability hazard causes a fatal injury*.

*Statistical distributions* were used to define the NPTs for some continuous (numeric) variables based on their purpose. For instance, continuous variables without parents are defined using a Uniform distribution or using a TNormal distribution given prior data for a medical device. Continuous variables with parents are defined using a TNormal distribution with mean $\mu$ as a weighted function of its parents and variance $\sigma^2$ or using a Binomial distribution, $B(n,p)$. *Comparative expressions* were used to define the NPTs for discrete variables with binary states. For instance, the NPT for the variable *Fatal injury risk acceptability* is defined using the following IF statement: *if (prob. of fatal injury per demand <= acceptable fatal injury probability, "Acceptable", "Not Acceptable")*.

Please note that the NPTs used for some of the nodes (variables) in the model will be revised, given the data and requirements for a particular medical device. In this paper, NPTs with statistical distributions include a sufficiently large variance to make them applicable to the different risk management scenarios presented in Section 5.

**4.3 The hybrid BN for medical device risk assessment and management**

The schematic shown in Figure 4 describes the hybrid Bayesian network (BN) for medical device risk assessment and management (see Figures 5a and 5b for BN model components).

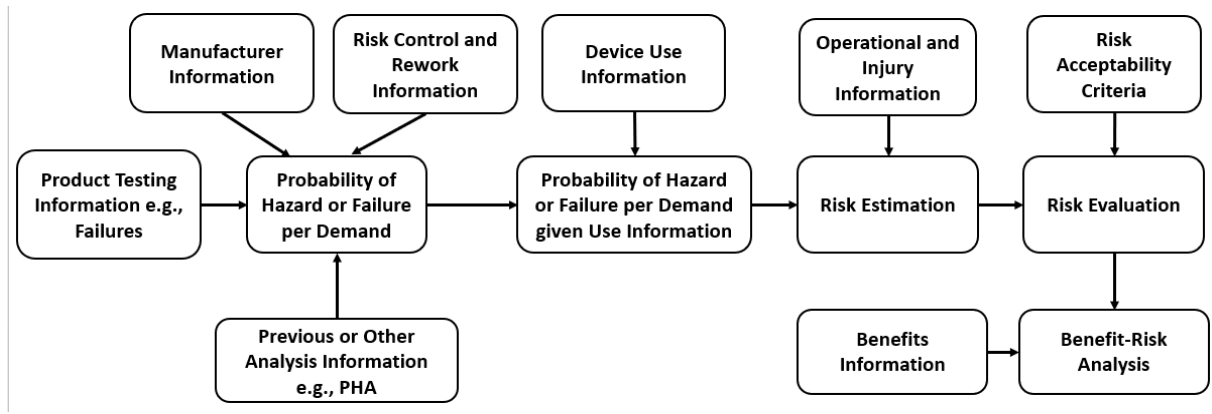

Figure 4. Schematic view of BN for medical device risk assessment and management

The following steps describe the procedure for using the BN for medical device risk assessment and management:

1. Define the scope and objectives of the analysis, including hazards to be investigated and risk acceptability criteria.
2. Describe the device, including its requirements, functions, users, intended use, safety characteristics, benefits, risk controls and life cycle phase.
3. Collate and organise other relevant information for the analysis:
    a. Product testing information: Information about the number of hazards observed in a set of demands during testing will allow the BN to estimate the probability of the hazard per demand. We define a *demand* as a measure of usage, e.g., single use, years etc.
    b. Injury information: Information about hazard occurrences and related injuries in the field will allow the BN to estimate the probability of the hazard or hazardous situation resulting in injury. Injury information can be obtained from hospitals and injury databases.
    c. Manufacturer information: Information such as manufacturer reputation, customer satisfaction, and product defects will allow the BN to estimate the manufacturing process quality. Since the manufacturing process quality can

influence the occurrence of hazards, it will be used to revise the probability of the hazard per demand, especially in situations where there is little or no product testing data.

4. Perform the analysis using the BN:
    a. Populate product testing information, manufacturer information, injury information and risk acceptability criteria.
    b. Compute the risk and overall residual risk acceptability.
    c. Estimate the effect of additional risk controls: In situations where the overall residual risk is not acceptable, populate risk control and rework information to estimate the residual risk given additional risk controls.
    d. Perform benefit-risk analysis: Populate benefits information to determine if the risk of the device is acceptable given its benefits. This is useful, especially in situations where the overall residual risk is not acceptable after additional risk controls are implemented or situations where risk control measures are not practicable.

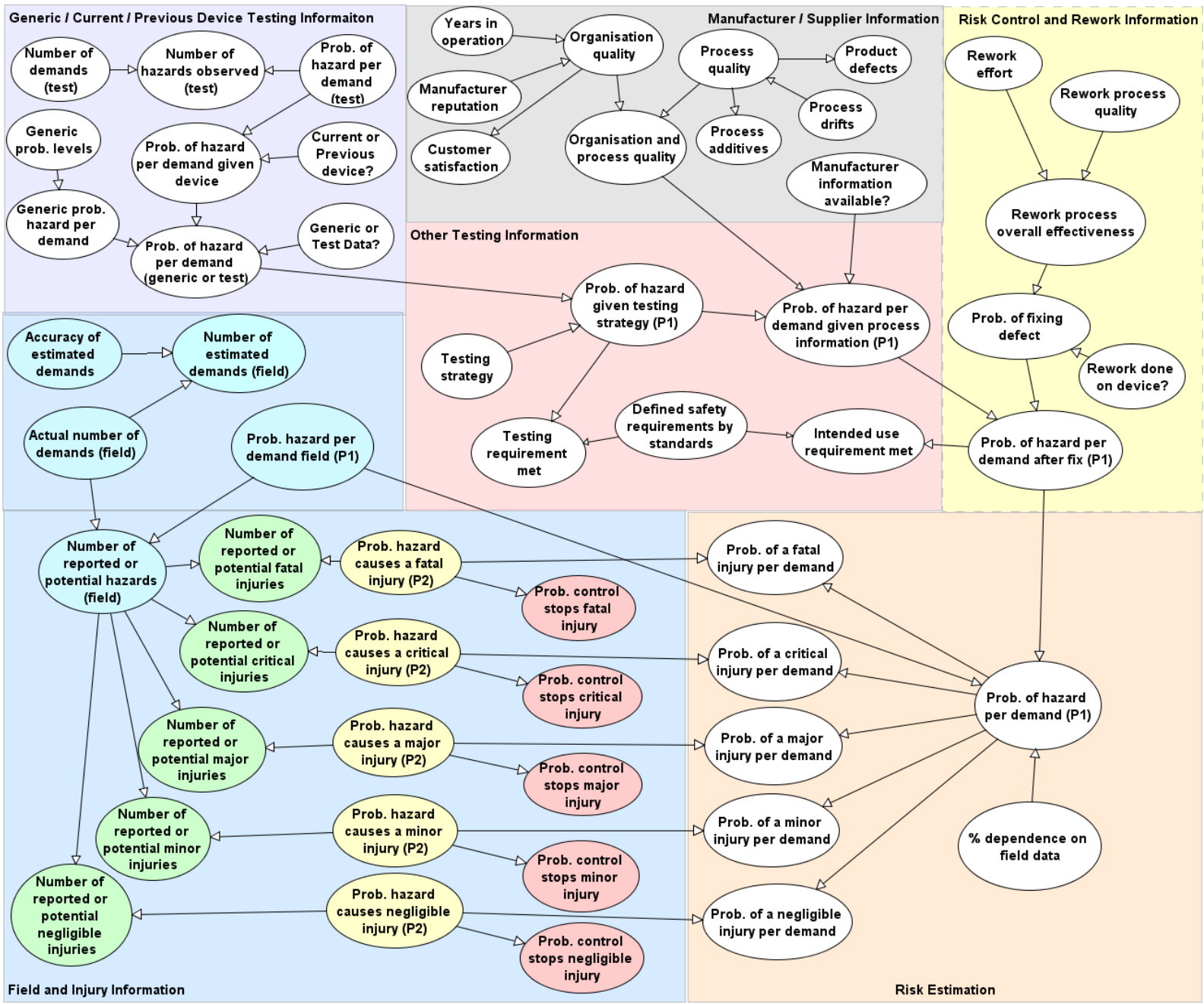

Figure 5a. Hybrid BN for medical device risk assessment and management – Component 1

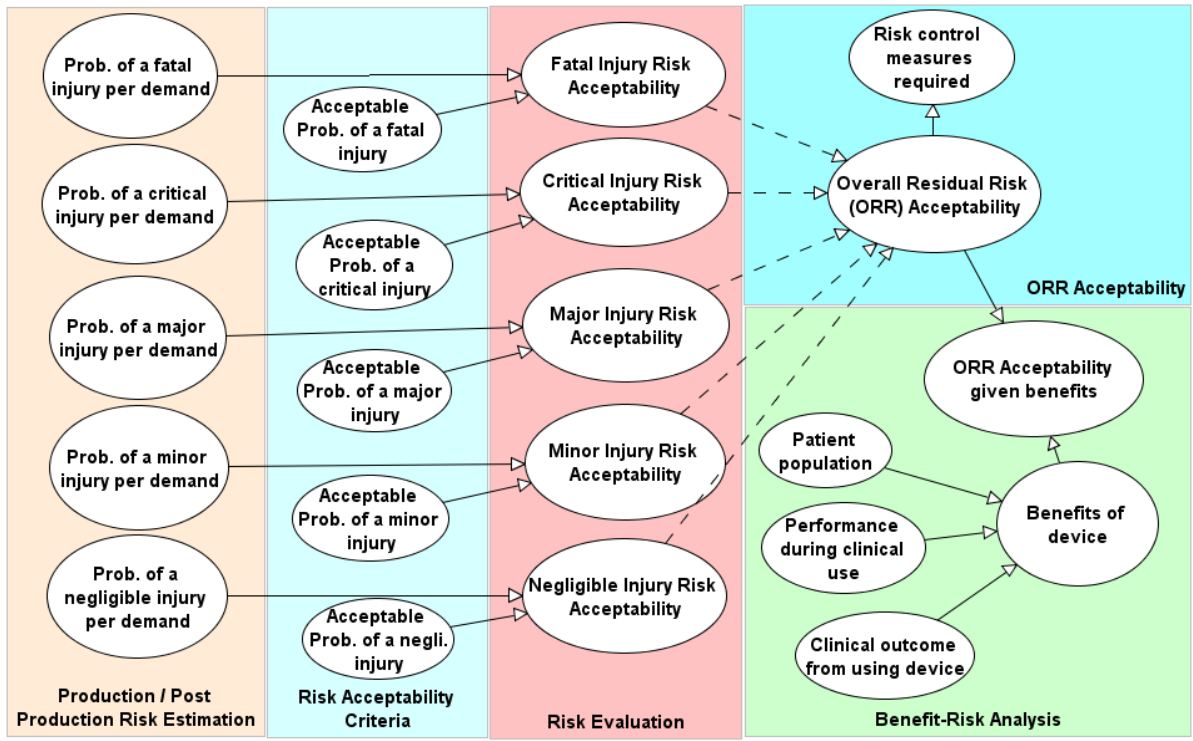

Figure 5b. Hybrid BN for medical device risk assessment and management – Component 2

Model assumptions include:

1. Five injury severity levels i.e., fatal, critical, major, minor and negligible. (See Supplemental Table 2)
2. An injury risk is judged acceptable if it is less than or equal to risk acceptability criteria.
3. There are hidden nodes whose parents are the 'Injury Risk Acceptability' nodes. These hidden nodes are used to translate the results of the discrete nodes used for 'Injury Risk Acceptability' to ranked nodes for computing the 'Overall Residual Risk Acceptability' (defined as a ranked node). This explains why there are dotted lines in the model.
4. The NPT for 'Overall Residual Risk Acceptability' node is defined as *TNormal (wmean (10.0, fatal injury acceptability, 4.0, critical injury acceptability, 3.0, major injury acceptability, 2.0, minor injury acceptability, 1.0 negligible injury acceptability), 0.001, 0,1).* We used a weighted mean function to combine the respective scores for injury risk acceptability to produce an overall residual risk acceptability score. The nodes with higher weights will have a greater impact on the overall residual risk acceptability score.
5. A single known type of hazard is investigated. In Section 6, we discuss combining the risk results of different hazards for a medical device.

# 5. Results

In this section, we demonstrate medical device risk management using the BN by assessing the risk of a generic Defibrillator during production and post-production. We show how the BN can assess the risk of medical devices with available testing data and with little or no testing data. In Section 5.1, we consider a range of hypothetical data scenarios, while in Section 5.2, we validate the method using publicly available real-world data from the LIFEPAK 1000 defibrillator.

## 5.1 Defibrillator Example

### 5.1.1 Background information

- *Product description*: An automated external defibrillator (AED) that sends an electric shock to the heart to treat serious heart arrhythmias, e.g., ventricular fibrillation. It is designed to be easy to use by trained users.
- *Hazard*: Incorrect shock advice.
- *Hazardous situation*: The defibrillator gave an incorrect shock advice leading to asystole.
- *Injury information*: Injuries range from normal sinus rhythm (negligible) to asystole (fatal).
- *Benefits information*: The patient population is 'very high'. Also, the performance expected during clinical use is 'high', and the clinical outcome expected from using the device is 'very high'.
- *Risk acceptability criteria*: We assume the following risk acceptability criteria presented in Table 2.

Table 2. Risk acceptability criteria for defibrillator

| Fatal Injury | Critical Injury | Major Injury | Minor Injury | Negligible Injury |
|---|---|---|---|---|
| 6.2E-5 | 9.9E-5 | 2.5E-4 | 7.6E-3 | 1.0E-2 |

- *Product testing information*: 5 hazards in 1000 demands. The defibrillator was tested 'typical of normal use'.
- *Rework information*: In situations where rework is needed (in this example, we assume that rework is needed if the value of the state 'Yes' for the node 'Risk control measures required' is > 10%), we assume that the manufacturer rework process quality and effort is 'very high'.

- *Manufacturer information:* The manufacturer has been in operation for over 20 years and has a very good safety record for medical devices. The manufacturer also has a 'high' customer satisfaction rating, and there are no product defects, no process additives and no process drifts.

- *Reported field and injury information:* The injury statistics for the defibrillator are based on data reported in a study analysing the performance of AEDs used in the Netherlands between January 2012 and December 2014 (Zijlstra, Bekkers, Hulleman, Beesems, & Koster, 2017). According to the study data, there were "1091 shock advices in 3310 analysis periods (demands). 44 of 1091 shock advices were incorrect. 15 incorrect shock advices were caused by device-related errors, and 28 were caused by operator-related errors". Injuries caused by device-related errors include 3 asystole, 5 narrow complex tachycardia, 4 bradycardia, 2 normal sinus rhythm and 1 multiple PVCs. Based on the injury severity classes used in the BN model (see Supplemental Table 2), we assume that asystole is a fatal injury, bradycardia, narrow complex tachycardia, and multiple PVCs are major injuries, and normal sinus rhythm is a negligible injury. Hence, we assume 3 fatal injuries, 0 critical injuries, 10 major injuries, 0 minor injuries and 2 negligible injuries given 15 incorrect shock advices.

### 5.1.2 Risk Management Scenarios

***Scenario 1 – Production risk management (with available testing data):*** In this scenario, we assess the risk of the defibrillator given the information in Section 5.1.1.

***Scenario 2 – Production risk management (with limited or no testing data):*** In this scenario, we assume that the defibrillator is a novel device with no testing data. We assume that we have testing data from a previous similar defibrillator (5 hazards in 700 demands). We also assume that the P1 estimate (i.e., probability of hazard per demand) for the novel device is also dependent on P1 estimated from field data (ratio 60:40 i.e., $P1 = (0.60 \times P1\ test) + (0.40 \times P1\ field)$). All other information used in the model is stated in Section 5.1.1.

***Scenario 3 – Production risk management (generic data):*** In this scenario, we assume that the defibrillator is a novel device with no testing data. We assume we are using generic probabilities for the hazard occurrence (see Supplemental Table 3). We assume that the hazard occurrence is probable (i.e., 1E-4 ≤ P1 < 1E-3), and the P1 estimate for the novel device is also dependent on P1 estimated from field data (ratio 60:40). All other information used in the model is stated in Section 5.1.1.

*Scenario 4 – Post-production risk management*: In this scenario, we assume that we are reassessing the risk of a previous model of the defibrillator available on the market based on reported hazards and injuries. We assume 10,000 demands, 50 reports of incorrect shock advices resulting in 1 major injury and 49 negligible injuries. The risk acceptability criteria and benefits information used in the model is stated in Section 5.1.1.

### 5.1.3 Risk Management Results

*Scenario 1 - Production risk management (with available testing data)*: The BN risk results for the defibrillator are summarised in Table 3 (see Supplemental Figures 2a and 2b).

Table 3. Defibrillator (with available testing data) BN risk results

| Product | Defibrillator | | | | |
|---|---|---|---|---|---|
| **Injury Severity** | Fatal | Critical | Major | Minor | Negligible |
| **Risk Acceptability Criteria** | 6.2E-5 | 9.9E-5 | 2.5E-4 | 7.6E-3 | 1.0E-2 |
| **Risk Estimate (Median)** | 1.1E-3 | 2.07E-4 | 3.25E-3 | 2.07E-4 | 8.0E-4 |
| **Risk Acceptability** | 0.13% | 28.8% | 0.07% | 100% | 100% |
| **Overall Residual Risk (ORR) Acceptability** | 14.2% | | | | |
| **Additional risk controls required (criteria 10%)** | 85.6% | | | | |
| **Benefit-Risk Analysis: ORR Acceptability** | 66.6% | | | | |

The BN model predicted that the median value of the risk distribution for fatal, critical and major injuries exceeded the risk acceptability criteria, and additional risk controls are required. The risk acceptability percentage indicates the percentage of the risk distribution that is acceptable. In this scenario, the BN predicted that 0.13%, 28.8% and 0.07% of the distribution for fatal, critical and major injuries, respectively, are acceptable. The ORR acceptability was 14.2%, and given the benefits of the device, it was 66.6%. Please note that though there are no available data for critical and minor injuries, the BN model provides reasonable probabilities estimates based on the number of reported hazards and other evidence in the model.

In Table 4 (see Supplemental Figures 2c and 2d), we show the risk results if additional risk controls are implemented. As stated in Section 5.1.1, the manufacturer has a 'very high' quality rework process and effort.

Table 4. Defibrillator (with available testing data) BN risk results – Rework Information

| Product | Defibrillator | | | | |
|---|---|---|---|---|---|
| **Injury Severity** | Fatal | Critical | Major | Minor | Negligible |
| **Risk Acceptability Criteria** | 6.2E-5 | 9.9E-5 | 2.5E-4 | 7.6E-3 | 1.0E-2 |
| **Risk Estimate (Median)** | 2.2E-4 | 4.2E-5 | 6.6E-4 | 4.2E-5 | 1.6E-4 |
| **Risk Acceptability** | 4.5% | 78.1% | 4.4% | 100% | 100% |
| **Overall Residual Risk (ORR) Acceptability** | 28.7% | | | | |
| **Additional risk controls required (criteria 10%)** | 71.3% | | | | |
| **Benefit-Risk Analysis: ORR Acceptability** | 72.3% | | | | |

Compared to the results presented in Table 3, the BN model revised the risk estimates given additional risk controls. The results show that risk acceptability for fatal, critical and major injuries would increase to 4.5%, 78.1% and 4.4%, respectively. ORR would increase to 28.7%, and given benefits would increase to 72.3%.

*Scenario 2 - Production risk management (with limited or no testing data)*: The BN risk results for the defibrillator are summarised in Table 5 (see Supplemental Figures 3a and 3b).

Table 5. Defibrillator (with limited or no testing data) BN risk results

| Product | Defibrillator | | | | |
|---|---|---|---|---|---|
| **Injury Severity** | Fatal | Critical | Major | Minor | Negligible |
| **Risk Acceptability Criteria** | 6.2E-5 | 9.9E-5 | 2.5E-4 | 7.6E-3 | 1.0E-2 |
| **Risk Estimate (Median)** | 1.4E-3 | 2.6E-4 | 4.0E-3 | 2.6E-4 | 1.0E-3 |
| **Risk Acceptability** | 0.03% | 22.9% | 0% | 100% | 100% |
| **Overall Residual Risk (ORR) Acceptability** | 12.8% | | | | |
| **Additional risk controls required (criteria 10%)** | 87.2% | | | | |
| **Benefit-Risk Analysis: ORR Acceptability** | 66% | | | | |

*Scenario 3 - Production risk management (with generic data)*: The BN risk results for the defibrillator are summarised in Table 6 (see Supplemental Figures 4a and 4b).

Table 6. Defibrillator (with generic data) BN risk results

| Product | Defibrillator | | | | |
|---|---|---|---|---|---|
| **Injury Severity** | Fatal | Critical | Major | Minor | Negligible |
| **Risk Acceptability Criteria** | 6.2E-5 | 9.9E-5 | 2.5E-4 | 7.6E-3 | 1.0E-2 |
| **Risk Estimate (Median)** | 4.8E-4 | 9.2E-5 | 1.4E-3 | 9.2E-5 | 3.5E-4 |
| **Risk Acceptability** | 0.24% | 52.6% | 0% | 100% | 100% |
| **Overall Residual Risk (ORR) Acceptability** | 19.6% | | | | |
| **Additional risk controls required (criteria 10%)** | 80.4% | | | | |
| **Benefit-Risk Analysis: ORR Acceptability** | 68.7% | | | | |

*Scenario 4 – Post-production risk management*: The BN risk results for the defibrillator post-production are summarised in Table 7 (see Supplemental Figures 5a and 5b).

Table 7. Defibrillator (post-production) BN risk results

| Product | Defibrillator | | | | |
|---|---|---|---|---|---|
| **Injury Severity** | Fatal | Critical | Major | Minor | Negligible |
| **Risk Acceptability Criteria** | 6.2E-5 | 9.9E-5 | 2.5E-4 | 7.6E-3 | 1.0E-2 |
| **Risk Estimate (Median)** | 6.8E-5 | 6.8E-5 | 1.6E-4 | 6.8E-5 | 4.3E-3 |
| **Risk Acceptability** | 47% | 63% | 72% | 100% | 100% |
| **Overall Residual Risk (ORR) Acceptability** | 62% | | | | |
| **Additional risk controls required (criteria 10%)** | 38% | | | | |
| **Benefit-Risk Analysis: ORR Acceptability** | 85% | | | | |

### 5.2 Results Validation

To validate the model results, we assessed the risk of LIFEPAK 1000 Defibrillator (Product Part Numbers: 320371500XX), which was recalled by Physio-Control in 2017 due to reports of the device shutting down unexpectedly during device use (FDA, 2017, 2018a). This hazard can cause the device not to deliver therapy during use, exposing the patient to serious harm or death. A total of 133,330 devices were affected by this hazard. There were 34 reports of the hazard and 8 adverse events. In this example, we assume the risk acceptability criteria and benefits information stated in Section 5.1.1 since this information is not publicly available. We also assume that number of potentially fatal injuries is 133,330 since the hazard affected all product instances.

According to the model results shown in Table 8 (see Supplemental Figures 6a and 6b), the Fatal injury risk is unacceptable, and the overall residual risk is only 50% acceptable. Hence the BN model validates and supports Physio-Control product recall decision.

Table 8. BN model risk results for LIFEPAK 1000 Defibrillator

| Product | LIFEPAK 1000 Defibrillator | | | | |
|---|---|---|---|---|---|
| Injury Severity | Fatal | Critical | Major | Minor | Negligible |
| Risk Acceptability Criteria | 6.2E-5 | 9.9E-5 | 2.5E-4 | 7.6E-3 | 1.0E-2 |
| Risk Estimate (Median) | 0.9999 | 5.2E-6 | 5.2E-6 | 5.2E-6 | 5.2E-6 |
| Risk Acceptability | 0% | 99.9% | 100% | 100% | 100% |
| Overall Residual Risk Acceptability | 50% | | | | |
| Additional risk controls required (criteria 10%) | 50% | | | | |
| Benefit-Risk Analysis: ORR Acceptability | 81% | | | | |

## 6. Discussion

The Defibrillator example shows that the BN model can estimate the risk of medical devices during production and post-production with available relevant data and with limited or no relevant data. In Scenario 1 - production risk management, the BN model estimated the risk and acceptability for the defibrillator given relevant information (see Tables 3 and 4). In Scenario 2 - production risk management (with limited or no testing data), the BN model estimated the risk and acceptability for the defibrillator given limited product testing data using manufacturer information and previous similar device data (see Table 5). In Scenario 3 - production risk management (generic data), the BN model estimated the risk and acceptability for the defibrillator using generic probabilities of hazard occurrence along with manufacturer information and field data from other similar devices (see Table 6). In Scenario 4 - post-production risk management and model validation using LIFEPAK 1000 defibrillator data, the BN model estimated the risk and acceptability of the defibrillator based on operational and injury information (see Tables 7 and 8). In all scenarios, the risk estimate is comprehensive since the BN incorporates relevant factors that affect the risk of medical devices, such as the manufacturing process quality. Moreover, these factors are linked causally, supporting ease of interpretability and explanation of risk estimates. In fact, the BN model incorporates both discrete and continuous variables to estimate risk illustrating the flexibility and power of using hybrid BNs for solving complex problems. The BN uses continuous variables with conditionally deterministic functions, statistical distributions and mixture distributions conditioned on different discrete assumptions. Moreover, the BN can estimate risks using prior assumptions and learn parameters from observations (induction). Since the BN can easily revise risk estimates given new information, this allows easy risk management of any medical device throughout its life cycle.

The BN model also performs Benefit-Risk analysis by estimating the risk acceptability given the benefits of the medical device (see Figure 5b). The benefit of a medical device is the extent of improvement in a patient's health and clinical management expected from using that device. As shown in Figure 5b, information such as device performance and clinical outcomes can assist in determining the benefit of a medical device (ISO, 2020). A Benefit-Risk analysis is essential for informing risk management decisions such as product recalls, especially in situations where additional risk control measures are not applicable.

In situations where there is little or inadequate data to provide reasonable risk estimates for medical devices, the BN can incorporate data from previous similar devices, expert judgement, and manufacturer information to estimate the risk of the medical device. Previous similar device data, expert judgement and manufacturer information can be included as prior distributions or values in the BN, as illustrated in Scenario 2 and Scenario 3. Hence, the BN model can estimate the risk of novel medical devices (i.e., devices with little or no historical data) with known or unknown hazards or faults since it can handle uncertainty and incomplete data, combine subjective and objective evidence, and revise risk estimates given new evidence. In situations where the BN is used to assess the risk of a continuous use medical device, the BN can estimate the failure rate by considering the mean and variance of the observed failure times (demands) and the mean and variance of the number of observed failures.

In situations where the BN is used to assess the risk of software, information such as development team experience is required to determine the quality of the software development process. The BN can be adapted using the Software BN fragment shown in Figure 6 to estimate the quality of the software development process. The quality of the software development process is then combined with software failure data to provide a more accurate estimate of the probability of a software failure. Like novel medical devices, the BN can provide reasonable risk estimates for new software with little or no testing data by combining previous similar software failure data, expert judgement and knowledge about the development process. The estimated risk for the new software will be revised, given new evidence throughout its life cycle, such as rework information (risk control measure) and injury information.

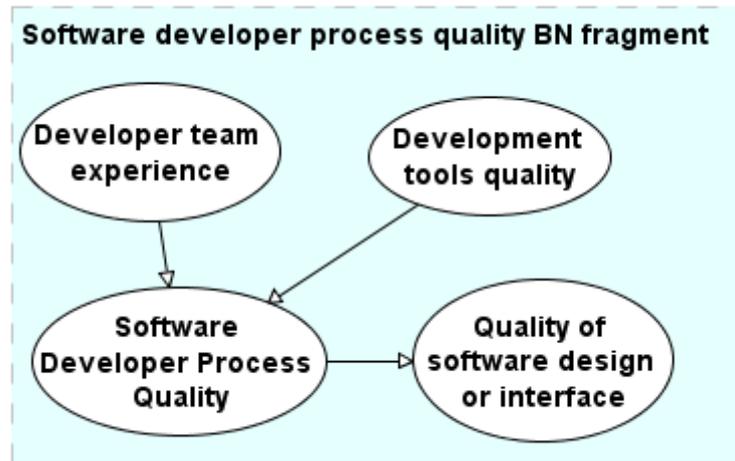

Figure 6. Software developer process quality BN fragment

In situations where the BN model is used to assess individual risk, the model can be extended to include information such as device use to estimate risk for a particular user. The BN model revises the P1 estimated from field data using device use information for that particular user. The revised P1 estimate is then combined with P2 information from field data to estimate injury risk as shown in Figure 7.

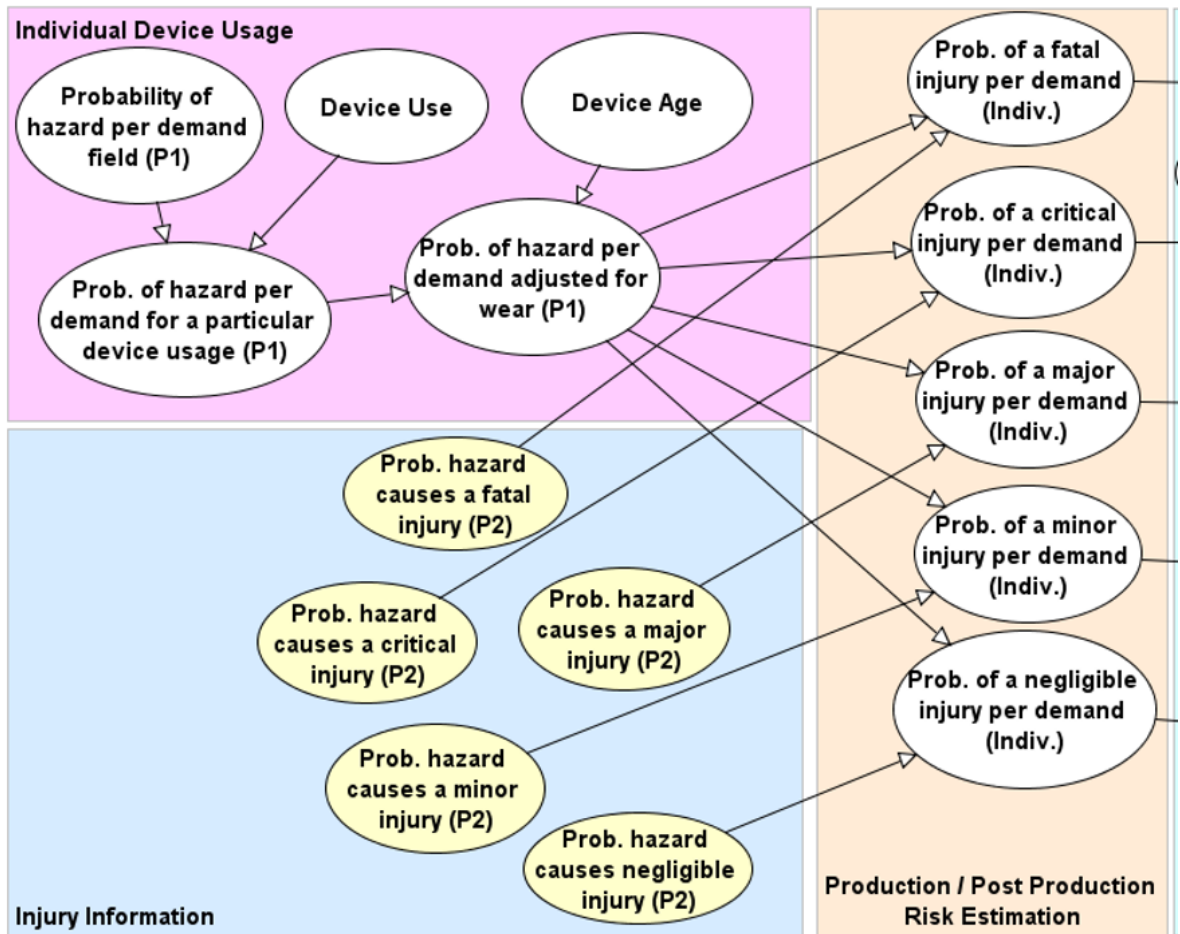

Figure 7. Individual Device Use Information BN fragment

The BN for medical device risk management provides risk estimates for a single hazard; however medical devices usually have multiple hazards. We can combine the results of multiple hazards using a matrix or table. In the example shown in Table 9, the acceptability value for each risk class for a particular hazard is obtained from the model, and we compute the combined acceptability values as the mean $\mu$ for each risk class. We assume that all values included in the table are satisfactory. The risk acceptability table will allow risk assessors to determine the overall risk acceptability for the medical device given all its hazards.

Table 9. Risk Acceptability Table

| Risk Acceptability Table | | | | | | | |
|---|---|---|---|---|---|---|---|
| **Risk Class** | Fatal | Critical | Major | Minor | Negligible | Overall residual risk | Overall residual risk given benefits |
| **Hazard 1** | 89% | 60% | 80% | 25% | 30% | 67% | 85% |
| **Hazard 2** | 50% | 75% | 100% | 99% | 75% | 75% | 90% |
| **Combined** | 70% | 68% | 90% | 62% | 53% | 71% | 87.5% |

Finally, we believe that the BN model improves medical device risk management by:

1. Providing a robust method for managing the risk of medical devices throughout their life cycle.
2. Informing risk control measures given risk acceptability criteria.
3. Improving interpretability and explanation of risk estimates.
4. Handling uncertainty in the data, especially for novel medical devices and software with little or no relevant historical data.
5. Incorporating device use and device age information when estimating individual risk.
6. Revising risk estimates given new information such as reported injuries.
7. Complementing existing risk management techniques and methods such as the FTA, PHA and the BXM method (Elahi, 2022).

Other benefits to using BNs that were not covered in this paper include *interventional* and *counterfactual* reasonings. Interventional reasoning entails predicting the effect of an intervention. It does not support diagnostic inference since the intervened variable is made independent of its causes via 'graph surgery'. Counterfactual reasoning entails investigating different outcomes had the observed events been different. It combines both observational and interventional reasoning using a twin-network model. A twin network model consists of two identical BNs representing real and counterfactual worlds linked via shared exogenous variables. However, in the counterfactual world where interventional reasoning is done, the intervened variable is independent of its parents (Balke & Pearl, 1994; Pearl, 2009; Pearl & Mackenzie, 2018).

## 7. Limitations

The main limitation of the study is obtaining all relevant information for a medical device to perform risk assessment using the BN model. Since the results of safety and reliability tests by manufacturers are not publicly available, some of the data used to assess the risk of the defibrillator were fictious such as the risk acceptability criteria. Given actual data for medical devices such as LIFEPAK 1000, the BN can provide reasonable, auditable risk estimates.

## 8. Conclusion and Recommendations

The proposed hybrid BN model for medical device risk management provides a systematic method for assessing the risk of medical devices. It can handle uncertainty, incomplete data, estimate risks using prior assumptions and learn parameters from observations (induction). It supports comprehensive and practical risk analysis since it decomposes the risk of a medical device into a causal chain of events, including controls and triggers, unlike the classical approach, i.e., $Risk = P \times S$. The BN model also complements existing medical device risk management tools and methods such as FTA and the BXM method (Elahi, 2022). The results of other risk assessment methods can be incorporated into the model to determine the risk of medical devices. Additionally, the BN model informs risk management decisions by providing information on risk acceptability and benefit-risk analysis.

We believe that the BN model can improve medical device risk management since it resolves the limitations of existing methods, provides a standard systematic method for medical device risk management, is generalisable, considers the ISO 14971 risk management process and complements existing risk analysis methods. Future work includes investigating risk perception of medical devices.

## Acknowledgements

We thank Mr. Bijan Elahi and Dr. Eric Maass for their useful comments and suggestions. This work was supported by the UK Government Department for Business, Energy and Industrial Strategy, Office for Product Safety and Standards (OPSS) and Agena Ltd. The views expressed in this article are those of the authors exclusively, and not necessarily those of the UK Government Department for Business, Energy and Industrial Strategy, Office for Product Safety and Standards (OPSS).